\def\year{2019}\relax
\documentclass[letterpaper]{article} 
\usepackage{aaai19}  
\usepackage{times}  
\usepackage{helvet}  
\usepackage{courier}  
\usepackage{url}  
\usepackage{graphicx}  
\usepackage{amsmath}
\usepackage{amssymb}
\usepackage{booktabs} 
\usepackage{caption}
\newcommand*\samethanks[1][\value{footnote}]{\footnotemark[#1]}
\begin{document}
%
\title{Unsupervised Stylish Image Description Generation via Domain Layer Norm}
\author{Cheng-Kuan Chen$\dagger$\thanks{Equal contribution},
        Zhu Feng Pan$\dagger$\samethanks,
        Min Sun$\dagger$,
        Ming-Yu Liu$\ddagger$\\
National Tsing-Hua University$\dagger$, Nvidia$\ddagger$
}
\maketitle

\begin{abstract}
Most of the existing works on image description focus on generating expressive descriptions. The only few works that are dedicated to generating \emph{stylish} (e.g.,\ romantic, lyric, etc.) descriptions suffer from limited style variation and content digression. To address these limitations, we propose a controllable stylish image description generation model. It can learn to generate stylish image descriptions that are more related to image content and can be trained with the arbitrary monolingual corpus without collecting new paired image and stylish descriptions. Moreover, it enables users to generate various stylish descriptions by plugging in style-specific parameters to include new styles into the existing model. We achieve this capability via a novel layer normalization layer design, which we will refer to as the Domain Layer Norm (DLN). Extensive experimental validation and user study on various stylish image description generation tasks are conducted to show the competitive advantages of the proposed model.
\end{abstract}

\section{Introduction}
\begin{figure}[t!]
    \centering
    \includegraphics[scale=0.26]{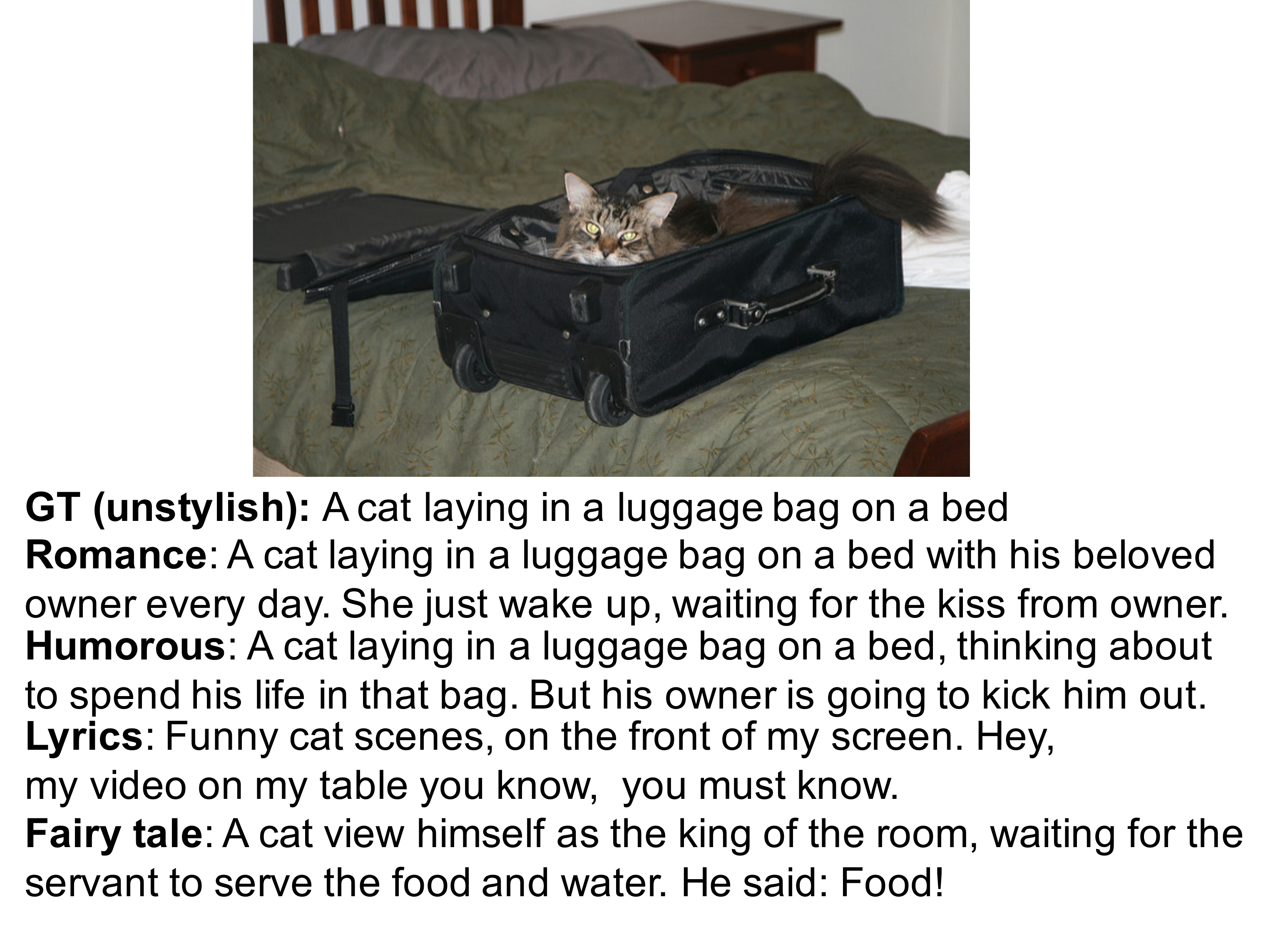}
    \caption{An ideal IDG can generate stylish descriptions for the given image. The generated descriptions should relate to the image content with different language styles.}
    \label{fig:Introduction}
\end{figure}

The image description generation (IDG) problem concerns about generating a natural language description that transcribes an input image. Over the years, tremendous effort has been dedicated to developing models that are descriptive. However, little effort is dedicated to generating descriptions that are \emph{stylish} (e.g.\ romantic, lyric, etc). Even for the handful of stylish IDG models that exist, they only have a loose control over the style. Ideally, a stylish IDG model should allow users to flexibly control over the generated descriptions as shown in Fig \ref{fig:Introduction}. Such a model would be useful for increasing user engagement in applications requiring human interaction such as chatbot and social media sharing. 

A naive approach to tackle the stylish IDG problem is to collect new corpora of paired images and descriptions for training. However, this is expensive. For each style that we wish to generate, we have to ask human annotators to write the romantic descriptions for each image in the training dataset. 

In this paper, we propose a controllable stylish IDG model. Our model is jointly trained with a paired unstylish image description corpus (source domain) and a monolingual corpus of the specific style (target domain). In this setting, our model can learn to generate various styles without collecting new paired data in the target domain. Our main contribution is to show that the layer normalization can be used to disentangle language styles from the content of source and target domains via a small tweak. This design enables us to use the shared content to generate descriptions that are more relevant to the image as well as control the style by plugging in a set of style-specific parameters. We refer this mechanism as Domain Layer Normalization (DLN) since we treat each style as the target domain in the domain transfer setting.

We conduct an extensive experimental evaluation to validate the proposed approach using both subjective and objective performance metrics. We evaluate our model on four different styles, including fairy tale, romance, humor, and country song lyrics style (lyrics). Experiment results show that our model generates stylish descriptions that are more preferred by human subjects. It also outperforms prior works on the objective performance metrics.
\section{Related Works}
\noindent \textbf{Visual style transfer.} Image style transfer has been widely studied in computer vision. Gatys et al.~\cite{gatys2015neural} synthesize a new stylish image by recombining image content with style features extracted from different images. Dumoulin et al.~\cite{45832} propose to learn the style embedding of visual artistic style by conditioning on the parameter of batch normalization~\cite{43442}. Huang et al.~\cite{huang2017arbitrary} use adaptive instance norm. More recent approaches use the generative adversarial network (GAN)~\cite{NIPS2014_5423} to align and transfer images from different domains. Liu et al.~\cite{NIPS2016_6544} employ weight-sharing assumption to learn the shared latent code between two domains and further propose translation stream in~\cite{NIPS2017_6672} to encourage the same image in two domains to be mapped into common latent code. While our method is similar to these works in high level, the discrete property of language required new model design.

\noindent \textbf{Language style transfer.} Supervised learning can be used to generate various linguistic attribute (e.g., different sentiments and different degrees of descriptiveness), but it requires a significant amount of labeled data. Many recent works assume there exist a share content space and a latent style vector between two non-parallel corpora for unsupervised language style transfer. Shen et al.~\cite{shen2017style} propose an encoder-decoder structure with adversarial training to learning this space. Following the same line, Melnyk et al.~\cite{melnyk2017improved} introduce content preservation loss and classification loss to improve the transfer performance. Fu et al.~\cite{fu2018style} propose to use a multi-decoder for different styles and a discriminator to learn a shared content code. Zhang et al.~\cite{zhang2018shaped} also use similar structure by using shared and private encoder-decoder. In a recent work, Prabhumoye et al.~\cite{prab2018style} introduce to ground the sentence in translation model, then apply adversarial training to get the desired style. What differs us from prior works is that we require generated stylish descriptions to match the visual content. Moreover, the style transferred in our work is more abstract instead of explicit styles such as sentiment, gender, or authorship in previous works.

\noindent\textbf{Image description generation.} Several works have been proposed to generate image descriptions by using paired image description data~\cite{vinyals2015show,krause2016paragraphs,liang2017recurrent}. To increase the naturalness and diversity of generated descriptions, Dai et al.~\cite{dai2017towards} apply adversarial training approach to train an evaluator to score the quality of generated descriptions. Chen et al.~\cite{chen2017show} propose an adversarial training procedure to adapt image captioning style using unpaired images and captions. A new objective is proposed in~\cite{dai2017contrastive} to enhance the distinctiveness of generated captions. On the other hand, there exist a few works proposed to enhance the attractiveness and style of the generated descriptions. Zhu et al.~\cite{zhu2015aligning} align the book and the corresponding movie release to a story-like description of the visual content. However, this method does not preserve the visual content. Matthews et al.~\cite{mathews2016senticap} propose the switch RNN to generate caption with positive and negative sentiments, which requires word level supervision and might not be able to scale. Recently, Gan et al.~\cite{gan2017semantic} investigate to generate tag-dependent caption by extending the weight matrix of LSTM to consider tag information. The following work StyleNet~\cite{gan2017stylenet} explores to decomposes LSTM matrix to incorporate the style information. One key difference is that we leverage an arbitrary stylish monolingual corpus that is not paired with any image dataset as target corpus instead of using paired images with stylish ground truth. The most similar to our work is~\cite{mathews2018semstyle}, the major differences are that we do not exploit the language features such as POS tag of corpus and we do not pre-process the target corpus to make it similar to the source one. Our approach is end to end with minimal pre-process of target corpus.

\section{Unsupervised Stylish Image Description Generation}
The goal of stylish Image Description Generation (IDG) is to generate a natural language description $d_T$ in space $\mathcal{D}_T$ given an image $I$ in the image space $\mathcal{I}$. The style of the description is implicitly captured in the description space $\mathcal{D}_T$, where we use subscript $T$ to emphasize the target style. There exist two settings for learning a stylish IDG model.
\paragraph{\bf Supervised stylish IDG.}  In supervised stylish IDG, we are given a training dataset $\mathbb{D}=\{(I^{(n)},d_T^{(n)}), n=1,...,N\}$, where each sample $(I^{(n)},d_T^{(n)})$ is a pair of image and its target stylish description sampled from the joint distribution $p(\mathcal{I},\mathcal{D}_T)$. The goal is to learn the conditional distribution $p(\mathcal{D}_T|\mathcal{I})$ using $\mathbb{D}$ so that we can generate \mbox{stylish image descriptions for an input image.}
\paragraph{\bf Unsupervised stylish IDG.}  In unsupervised stylish IDG, we are given two training datasets $\mathbb{D}_S$ and $\mathbb{D}_T$. $\mathbb{D}_S=\{(I^{(n)},d_S^{(n)}), n=1,...,N_S\}$ consists of pairs of image and its description $(I^{(n)},d_S^{(n)})$ sampled from $p(\mathcal{I},\mathcal{D}_S)$, where $S$ is referred to as the source domain which is typically unstylish. $\mathbb{D}_T=\{(d_T^{(n)}), n=1,...,N_T\}$ is a dataset of target stylish descriptions $d_T^{(n)}$ sampled from $p(\mathcal{D}_T)$, where the corresponding images are not available. Hence, the learning task is considered as unsupervised. The goal of unsupervised stylish IDG is to learn the conditional distribution $p(\mathcal{D}_T|\mathcal{I})$ using $\mathbb{D}_S$ and $\mathbb{D}_T$.

Unsupervised stylish IDG is an ill-posed problem since it is about learning the conditional distribution $p(\mathcal{D}_T|\mathcal{I})$ without using samples from the joint distribution $p(\mathcal{I},\mathcal{D}_T)$. Therefore, learning an unsupervised stylish IDG function is difficult without leveraging some useful assumptions. However, under the unsupervised setting, training data collection is greatly simplified: one could pair a general image description dataset (e.g., the MS-COCO dataset~\cite{lin2014microsoft}) with an existing corpus of the target style (e.g., some romantic novels) for learning. A solution to the unsupervised problem could enable many stylish image description generation applications.

\subsection{Unsupervised Stylish IDG via Domain Layer Norm}
\label{sec:framework}
\begin{figure*}[t]
    \centering
    \includegraphics[width=0.75\textwidth]{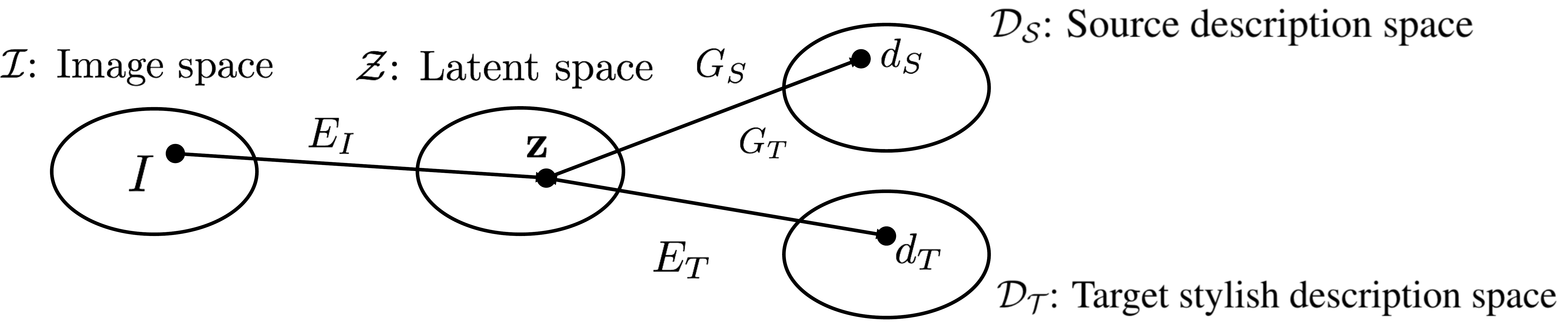}
    \caption{We make several assumptions to deal with the challenging unsupervised stylish image description generation problem. We first assume there exists a shared latent space $\mathcal{Z}$ so that a latent code $\mathbf{z}\in\mathcal{Z}$ can be mapped to the source description space $\mathcal{D}_S$ and the target stylish description space $\mathcal{D}_T$ via $G_S$ and $G_T$. We also assume there exists a stylish image description embedding function $E_T$ that can map a stylish description to a latent code. Finally, we assume there exists an image embedding function $E_I$ that can map an image to a latent code. 
    Once these functions are learned from data, we can generate a stylish image description for an image by applying $E_I$ and $G_T$ sequentially.
    }
    \label{fig::asssumption}
\end{figure*}
\paragraph{\bf Assumptions.}
To deal with the ill-posed unsupervised stylish IDG problem, we make several assumptions illustrated in Figure~\ref{fig::asssumption}. We first assume that there exists a latent space $\mathcal{Z}$ providing a common ground to effectively map to and from the image space $\mathcal{I}$, the source description space $\mathcal{D}_S$, and the target stylish description space $\mathcal{D}_T$. 
From latent space to description space, we assume that there exists a source description generation function $G_S(\mathbf{z})\in\mathcal{D}_S$ and a target stylish description generation function $G_T(\mathbf{z})\in\mathcal{D}_T$.
From non-latent space to latent pace, we assume that there exist an image encoder $E_I(I)\in\mathcal{Z}$ and a target description encoder $E_T(d_T)\in\mathcal{Z}$.
Our goal is to learn the generation functions ($G_T$ and $G_S$) and the encoding functions ($E_I$ and $E_T$) from the unsupervised stylish IDG training data $\mathbb{D}_S$ and $\mathbb{D}_T$. Note that this is a challenging learning task if $G_T$ and $G_S$ is completely independent of each other. Hence, we assume that $G_T$ and $G_S$ share the ability to describe the same factual content but with different styles.
Once these functions are learned,  we can simply first encode the image $I$ to a latent code using $E_I$ and then using $G_T$ to generate a stylish image description. In other words, the stylish image description is given by $G_T(E_I(I))$. We model the conditional distribution as $p(\mathcal{D}_T|\mathcal{I})=\delta(G_T(E_I(I)))$, where $\delta$ is the delta function.
\begin{figure*}[t]
    \centering
    \includegraphics[width=0.78\textwidth]{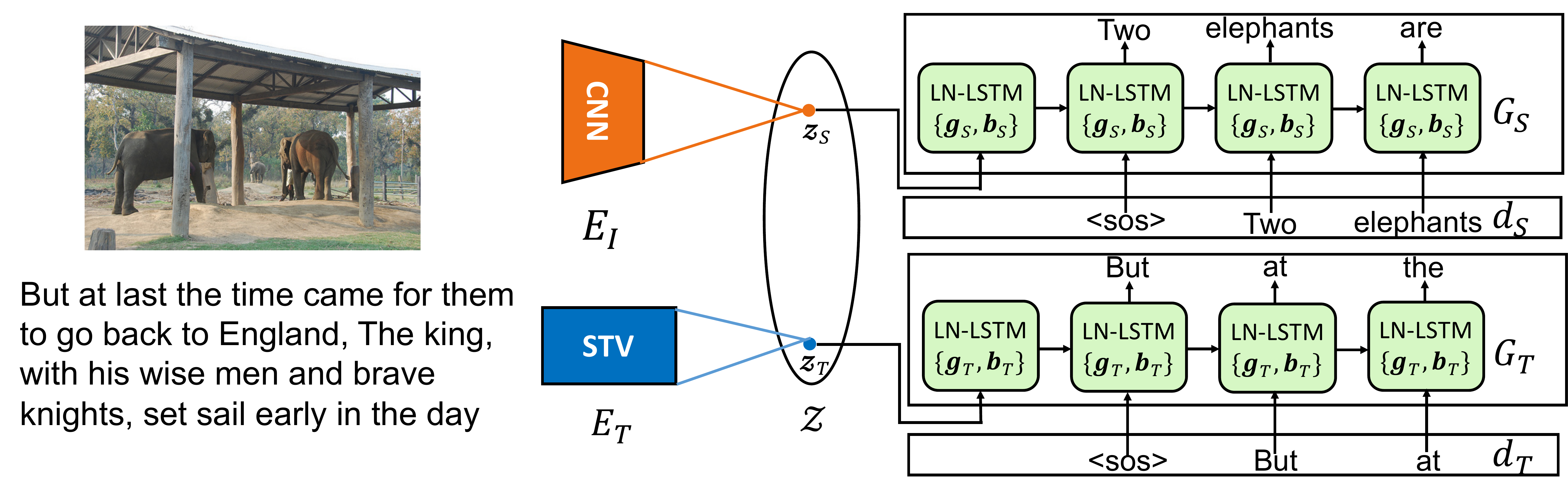}
    \caption{The $E_{I}$ and $E_{T}$ map the image and the target stylish description to a shared latent space. Both $G_{S}$ and $G_{T}$ share all weights except the layer norm parameters to capture the similar content in two domains. To disentangled the style factor, we employ different sets of layer norm parameters denoted as $\{\boldsymbol{g}_{S},\boldsymbol{b}_{S}\}$ and $\{\boldsymbol{g}_{T},\boldsymbol{b}_{T}\}$ for source and target domain during training.}
    \label{fig::model_archi}
\end{figure*}
Inspired by the success of deep learning, we model both of the generation and encoding functions using deep networks.
Specifically, we model $E_I$ using a deep convolutional neural network (CNN)~\cite{NIPS2012_4824} and model $E_T$, $G_T$, and $G_S$ using recurrent neural network as illustrated in Figure~\ref{fig::model_archi}. We also use Skip-Thought Vectors (STV)~\cite{kiros2015skip} to model $E_T$. For $G_T$ and $G_S$, we use Layer Normalized Long Short Term Memory unit (LN-LSTM) as their recurrent module~\cite{ba2016layer,hochreiter1997long}.
\begin{figure}[t]
    \centering
    \includegraphics[scale=0.18]{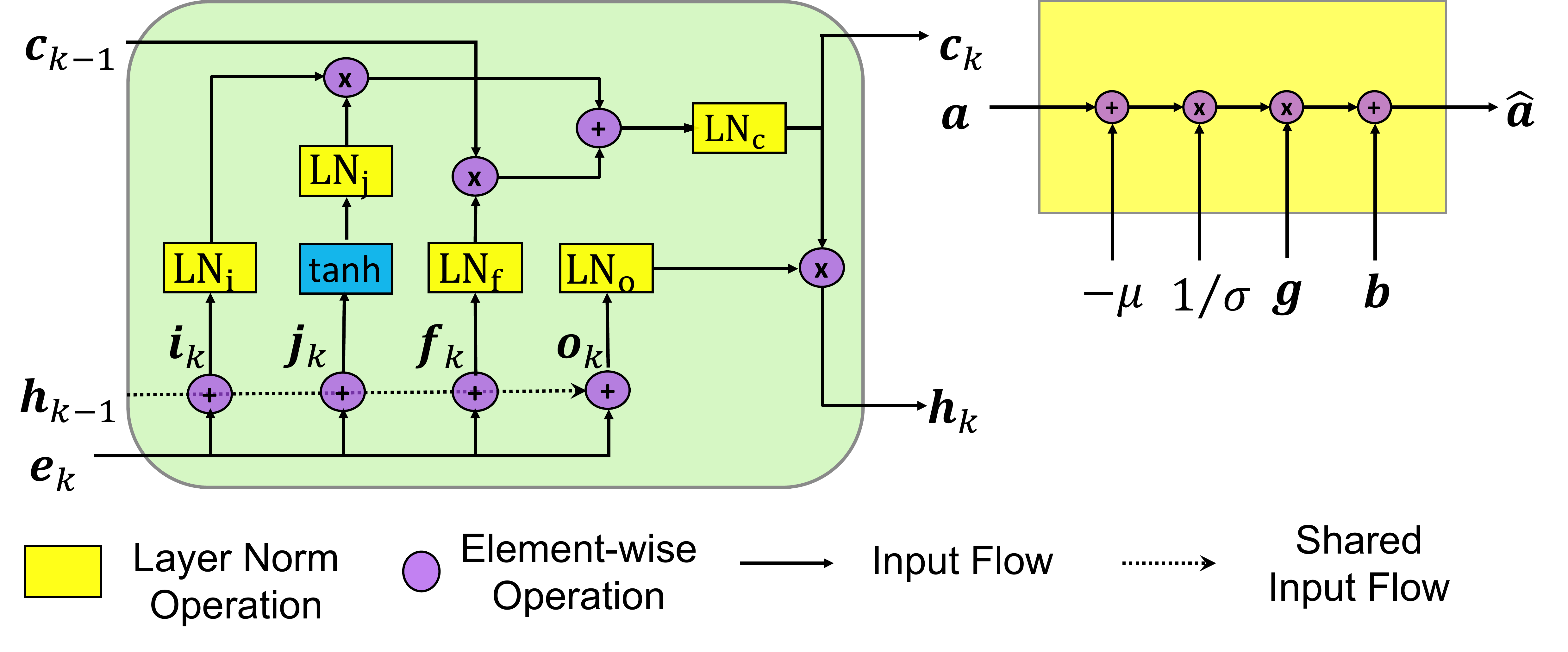}
    \caption{Inside the LN-LSTM cell (left) and the operation of layer normalization (right).}
    \label{fig:DLNdetail}
\end{figure}
\paragraph{\bf Training sketch.}
With the source domain dataset $\mathbb{D}_S$, we can train $\boldsymbol{z}_S=E_I(I)$ and $d_S=G_S(\boldsymbol{z}_S)$ jointly by solving the supervised IDG learning task, where $\boldsymbol{z}_S$ is the learned latent representation in the source domain. On the other hand, with the target domain dataset $\mathbb{D}_T$, we can train $\boldsymbol{z}_T=E_T(d_T)$ and $d_T=G_T(\boldsymbol{z}_T)$ jointly by solving an unsupervised description reconstruction learning task, where $\boldsymbol{z}_T$ is the learned latent representation in the target domain. To ensure that the latent space is shared (i.e., $\boldsymbol{z}_T\in \mathcal{Z}$ and $\boldsymbol{z}_S\in \mathcal{Z}$), we further assume that the generation functions $G_S$ and $G_T$ share most of their parameters. 

\paragraph{\bf Domain Layer Norm.}
Specifically, we assume $G_S$ and $G_T$ share all the parameters except those in their layer norm parameters~\cite{ba2016layer}. In other words, the domain description generators ($G_S$ and $G_T$) only defer in the layer norm parameters. We refer this weight-sharing scheme as the Domain Layer Norm (DLN) scheme. The intuition behind DLN is to encourage the shared weight to capture the factual content between two domains while the differences (i.e., styles) are captured in layer norm parameters. This design helps $G_T$ generate descriptions that are related to the image content even without the supervision of the corresponding images in training.

\noindent\textbf{Training $E_I$ and $G_S$ via Supervised IDG.}
The goal of supervised image description generation is to learn $p(\mathcal{D}_S|\mathcal{I})$ by using $\mathbb{D}_S$. The $G_S$ consists of an embedding matrix $\boldsymbol{\theta_W}$ that maps input text $x_k$ to a vector $\boldsymbol{e}_k$, an LN-LSTM module, and an output matrix $\boldsymbol{\theta_V}$ that maps hidden state to predicted token $\boldsymbol{\hat{y}}$. Formally,
\begin{align}\label{eq:1}
(\boldsymbol{\hat{y}}_{k+1}, \boldsymbol{h}_{k+1}) = G_S(\boldsymbol{e}_k,\boldsymbol{h}_{k})~,\\ \boldsymbol{\hat{y}}_{k+1} = \boldsymbol{\theta_V}^T\boldsymbol{h}_{k}~,\\
\boldsymbol{e}_k = \boldsymbol{\theta_W}^T\boldsymbol{1}\{x_k\}~,\\
\boldsymbol{e}_{-1}= E_I(I), \boldsymbol{h}_{-1} = \boldsymbol{0}~,
\end{align}
where $\boldsymbol{h}_{k}$ is the hidden feature in the LN-LSTM, $k\in\{-1\ldots m-1\}$ is time step of description with length $m$, and $\boldsymbol{1}\{\}$ denotes the operator for one-hot encoding. To train the network, we minimize the sum of cross-entropy of correct words as follows,
\begin{align}
    \mathcal{L}_{S} &= -\sum_{k=1}^{m}\textrm{log}(\boldsymbol{1}\{x_k\}^T\boldsymbol{\hat{y}}_k)~,
\end{align}
where $x_k$ is the $k^{th}$ word in the ground truth sentence.

\noindent\textbf{Training $E_T$ and $G_T$ via Stylish Image Description Reconstruction.}
The $G_T$ contains the LN-LSTM module, the same output matrix and embedding matrix used in $G_S$. Formally,
\begin{align}\label{eq:3} 
(\boldsymbol{\hat{y}}_{k+1}, \boldsymbol{h}_{k+1}) = G_T(\boldsymbol{e}_k,\boldsymbol{h}_{k})~,\\ \boldsymbol{\hat{y}}_{k+1} = \boldsymbol{\theta_V}^T\boldsymbol{h}_{k}~,\\
\boldsymbol{e}_k = \boldsymbol{\theta_W}^T\boldsymbol{1}\{d^k_T\}~,\\
\boldsymbol{e}_{-1} = E_T(d_T)~,\\
\boldsymbol{h}_{-1} = \boldsymbol{0}~,
\end{align}

where $d_T$ is the target style image description. To train the network, we minimize the reconstruction error as follows,
\begin{align}
    \mathcal{L}_{T} &= -\sum_{k=1}^{m}\textrm{log}(\boldsymbol{1}\{d^k_T\}^T\boldsymbol{\hat{y}}_k)~,
\end{align}
where $d^k_T$ is the $k^{th}$ word in the target style image description.

\noindent\textbf{Relating $G_S$ and $G_T$ via Domain Layer Norm.}
We relate $G_{S}$ and $G_{T}$ by sharing all weights except layer norm parameters in the LN-LSTM. Details inside the LN-LSTM are shown in Fig \ref{fig:DLNdetail}, where the layer norm operation (LN) is applied to each gate of LSTM. Take the input gate as an example:
\begin{align}\label{eq:4}
\boldsymbol{\hat{i}}_k &= \textrm{LN}(\boldsymbol{i}_k), \boldsymbol{i}_k = \boldsymbol{\theta}_{ie}\boldsymbol{e}_k+ \boldsymbol{\theta}_{ih}\boldsymbol{h}_{k-1}~,
\end{align}
where $\boldsymbol{\hat{i}}_k$ and $\boldsymbol{i}_k$ are the normalized and unnormalized input gates, $\boldsymbol{\theta}_{ie}$, $\boldsymbol{\theta}_{ih}$ are two projection matrices that map the embedding vector and the previous hidden state into the same dimension. The LN operation converts any input $\boldsymbol{a}$ to a normalized output $\boldsymbol{\hat{a}}$ as follows,
\begin{align}
\label{eq:5}
    \boldsymbol{\hat{a}} &= \frac{\boldsymbol{g}}{\sigma} \odot (\boldsymbol{a} - \mu)) + \boldsymbol{b}~,\\
    \mu &= \frac{1}{p_{h}}\sum_{i=1}^{i=p_{h}}a_{i}~,\\
    \sigma &= \sqrt{\frac{1}{p_{h}}\sum_{i=1}^{p_{h}}(a_{i} - \mu)^2}~,
\end{align}
where $a_i$ denotes the $i^{th}$ entry in the vector $\boldsymbol{a}$, $p_h$ is the dimention of the input $\boldsymbol{a}$, $\mu$ and $\sigma$ are the mean and standard deviation of the input $\boldsymbol{a}$, $\boldsymbol{g}$ and $\boldsymbol{b}$ are scaling and shifting vectors (i.e., layer norm parameters) learned from the data.

We train the whole network by jointly minimizing the supervised IDG loss $\mathcal{L}_{S}$ and the unsupervised image description reconstruction loss $\mathcal{L}_{T}$ subject to the architectural constraint set to $G_S$ and $G_T$ as below, where $\lambda$ is a hyperparameter.
\begin{multline}\label{eq:6}
    \mathcal{L}(\boldsymbol{\theta}_{E_I} \boldsymbol{\theta}_{G_S},\boldsymbol{\theta}_{E_T},\boldsymbol{\theta}_{G_T}) = \lambda \mathcal{L}_{S}(\boldsymbol{\theta}_{E_I},\boldsymbol{\theta}_{G_S})\\
    + (1 - \lambda)\mathcal{L}_{T}(\boldsymbol{\theta}_{E_T},\boldsymbol{\theta}_{G_T})~.
\end{multline}
\paragraph{\bf Extension to New Target Styles.}
Given a model with parameters $\boldsymbol{\theta_{V}}$, $\boldsymbol{\theta_{W}}$, $\boldsymbol{\theta}_{E_I}$, and $\boldsymbol{\theta}_{G_S}$, pre-trained on a pair of the source and one target domain, we aim to adapt it to a new target domain (i.e, style) by enlarging $\boldsymbol{\theta_{V}}$ and $\boldsymbol{\theta_{W}}$ to $\boldsymbol{\theta}^{\prime}_{\boldsymbol{V}}$ and $\boldsymbol{\theta}^{\prime}_{\boldsymbol{W}}$ to accommodate new vocabulary and finetuning the remaining parameters to $\boldsymbol{\theta}^{\prime}_{E_I}$, $\boldsymbol{\theta}^{\prime}_{E_T}$, $\boldsymbol{\theta}^{\prime}_{G_S}$ and $\boldsymbol{\theta}^{\prime}_{G_T}$. Hence, we define a new loss function as:
\begin{multline}\label{eq:7}
    \mathcal{L}(\boldsymbol{\theta}^{\prime}_{E_{I}}, \boldsymbol{\theta}^{\prime}_{G_S},\boldsymbol{\theta}^{\prime}_{E_T},\boldsymbol{\theta}^{\prime}_{G_T}) = \lambda_1\mathcal{L}_{S}(\boldsymbol{\theta}^{\prime}_{E_I},\boldsymbol{\theta}^{\prime}_{G_S})\\ + (1 -\lambda_1)\mathcal{L}_{T}(\boldsymbol{\theta}^{\prime}_{E_T},\boldsymbol{\theta}^{\prime}_{G_T}) + \lambda_2R(\boldsymbol{\theta}^{\prime}_{E_{I}},\boldsymbol{\theta}^{\prime}_{\boldsymbol{W}}, \boldsymbol{\theta}^{\prime}_{\boldsymbol{V}})~,
\end{multline}
where $\lambda_1$ and $\lambda_2$ are hyperparameters.
The regularization term $R(\boldsymbol{\theta}^{\prime}_{E_{I}},\boldsymbol{\theta}^{\prime}_{\boldsymbol{W}}, \boldsymbol{\theta}^{\prime}_{\boldsymbol{V}})=\lVert \boldsymbol{\theta}^{\prime}_{E_{I}} - \boldsymbol{\theta}_{E_{I}}\rVert_{2} + \lVert \boldsymbol{\theta}^{\prime}_{\boldsymbol{W}} - \boldsymbol{\theta_{W}} \rVert_{2} + \lVert \boldsymbol{\theta}^{\prime}_{\boldsymbol{V}} - \boldsymbol{\theta_{V}} \rVert_{2}$ is used to prevent new weights from deviating the pretrained model. This encourages the adapted model to keep the information learned during the pretrained phase. 
We use pretrained $\boldsymbol{\theta}_{E_I}$ and $\boldsymbol{\theta}_{G_S}$ as initialization of $\boldsymbol{\theta}^{\prime}_{E_I}$ and $\boldsymbol{\theta}^{\prime}_{G_S}$. For $\boldsymbol{\theta}^{\prime}_{G_T}$, we share all parameters in $\boldsymbol{\theta}^{\prime}_{G_S}$ except the layer norm parameters. $\boldsymbol{\theta}^{\prime}_{E_T}$ is trained from scratch. Note that we do not update the source domain layer norm parameters since we do not need to learn source style.
\section{Experiment}
We conduct two experiments to evaluate our proposed method. First, we demonstrate that our method can generate stylish descriptions based on paired image and unstylish description in the source domain and a stylish monolingual corpus that is not paired with any image dataset in the target domain. Then, we demonstrate the flexibility of our DLN to progressively include new styles one by one in the second experiment. The implementation details are in the supplementary. 
\subsection{Evaluation Setting}
\noindent\textbf{Datasets.} We use paragraphs released in~\cite{krause2016paragraphs} (VG-Para) as our source domain dataset. We do not use caption dataset such as MS-COCO because we found captions are less stylish when transfer to target style domain. We use pre-split data which contain 14575, 2489 and 2487 for training, validation and testing.  For target dataset, we use humor and romance novel collections in BookCorpus~\cite{zhu2015aligning}. We also collect country song lyrics and fairy tale to show that our method is effective on corpora with different syntactic structures and word usage. More details can be found in supplementary materials.\\
\noindent\textbf{Baselines.} We compare our method with four baselines: StyleNet~\cite{gan2017stylenet}, Neural Story Teller (NST)~\cite{kiros2015skip}, DLN-RNN and Random. Stylenet generates stylish descriptions in an end-to-end way but with paired image and stylish ground truth description. NST breaks down the task into two steps, which first generate unstylish captions then apply style shift techniques to generate stylish descriptions. DLN-RNN uses the same framework as DLN with only difference in using simple recurrent neural network. Random samples the the same number of nouns as that in the unstylished ground truth from the corresponding vocabulary of target domain. Although a concurrent work ~\cite{mathews2018semstyle} that attempts to solve similar task as ours, the major differences are we do not exploit linguistic features and pre-process the target corpus to facilitate the training. Moreover, it is not sure whether the concurrent work can be applied to other styles or even multiple styles as it only makes a step toward generating sentences with romantic style.\\
\noindent\textbf{Metrics of semantic relevance.} As there is no ground truth sentences for stylish image descriptions in unpaired setting, the conventional n-gram based metrics such as BLEU~\cite{papineni2002bleu}, METEOR~\cite{denkowski2014meteor} and CIDEr~\cite{vedantam2015cider} cannot be applied. It is also not suitable to calculate these metrics between stylish sentences and the unstylished ground truth because the goal of stylish description generation is to change the word usage while preserve certain semantic relevance between the stylish description and images.

We propose content similarity to evaluate the semantic relevance between generated stylish sentences and the unstylished ground truth. To calculate content similarity, we define $C_{S}$ as the set of nouns in the ground truth (source domain), and $C^{\prime}_{S}$ as the union between $C_{S}$ and synonyms for each noun in $C_{S}$, for the model may describe the same object with different words (e.g., cup and mug). Similar logic is applied to $C_{T}$ and $C^{\prime}_{T}$ in the generated description (target domain). We calculate: 
\begin{align}
\label{eq:8}
p &= \frac{|C_T \cap C^{\prime}_{S}|}{|C_T|} & r = \frac{|C_S \cap C^{\prime}_T|}{|C_S|},~
\end{align}
We take the f-score of the $p$ and $r$ as the content similarity score. The overall content similarity score is averaged over the testing data. This is because we assume stylish descriptions should at least contain objects which appear in the image. We also report SPICE~\cite{anderson2016spice} score, which calculate the f-score of semantic tuples between untylished ground truth and the generated stylish descriptions. The final score is average over all testing data.\\
\noindent\textbf{Metrics of stylishness.} We use transfer accuracy to evaluate the stylishness of our generated description. The transfer accuracy is widely used in language style transfer task~\cite{shen2017style,melnyk2017improved,fu2018style}. It measures how often do descriptions have labels of target style on test dataset based on a pre-trained style classifier. We follow the definition of transfer accuracy in~\cite{fu2018style}, which is
\begin{equation}
\mathcal{T} =
\begin{cases}
1 & \text{if $s > 0.5$} \\
0 & \text{if $s \leq 0.5$}
\end{cases}
\end{equation}
where $s$ is the output probability score of the classifier. We define $R_{T} = \frac{N_{vt}}{N_{vs}}$ as our transfer accuracy, which is the fraction of number of testing $N_{vs}$ data in source domain and number of testing data that correctly transfer description with target style $N_{vt}$. The final score is average over all testing data.\\
\noindent\textbf{Human evaluation.} The difficulty in generating stylish sentence in unpaired setting is to remain semantic relevance. Therefore, we conduct a human study on Amazon Mechanical Turk (AMT) independently for each methods to judge the semantic relevance between image and description. For each model, we randomly sample 100 images then generate stylish descriptions for each style. Two workers are asked to vote the semantic relevance with following prompt: Given an image and a paragraph from the book (Our stylish corpus), how well does the paragraph content relate to objects in the image. Workers are forced to vote from unrelated to related. The criteria for eligible workers are having at least 100 successful HITs with 70\% acceptance rate. The total number of HIT is 2400. For each HIT, the order of options is randomized. Workers are forced to vote and all responses are counted without aggregation.
\begin{table*}[t!]
  \centering
  \begin{tabular}{lc|ccccccccc}
    \toprule
    Model  & Data & CS & S & T & $p$ & $r$ & $n_{p}$ & $n_{r}$\\
    \midrule
     NST~\cite{kiros2015skip} & Lyrics& 0.037 & 0.016  & 100\% & 0.041 & 0.044 & 0.68 & 0.75 \\
     StyleNet~\cite{gan2017stylenet}&Lyrics& 0.033  & 0.014 & 100\%    &0.038 & 0.038 &0.57 &0.67 \\
     Random              & Lyrics & 0.008 & 0.002 & 55.2\% & 0.007 & 0.012 &0.13 &0.09  \\
     DLN-RNN                  & Lyrics &0.072  & 0.030  &100\%  & \textbf{0.101} & 0.069 &\textbf{1.65} &1.17  \\    
     DLN  &Lyrics &\textbf{0.083}& \textbf{0.033} &99.2\% & 0.080& \textbf{0.115} &1.25 & \textbf{1.92} \\
    \midrule
     NST~\cite{kiros2015skip} & Romance& 0.088 & 0.039 & 100\% & 0.087 & 0.113 &\textbf{1.57} &1.90\\
     StyleNet~\cite{gan2017stylenet}   &Romance& 0.012 & 0.005 & 100\% & 0.032 &0.001 & 0.11&0.14  \\
     Random              & Romance& 0.005& 0.002&  100\% &0.004 &0.001 &0.07 &0.05 \\
     DLN-RNN                  & Romance& 0.083& 0.034  &   94.3\%     & 0.078 & 0.125      &1.27 &0.71 \\
     DLN                      &Romance &\textbf{0.151} & \textbf{0.058} & 95.4\%& \textbf{0.193}& \textbf{0.148} &1.56 &\textbf{2.43}\\
    \midrule
     NST~\cite{kiros2015skip} & Humor& 0.103 & 0.041 & 99.7\%& 0.097&0.143 &2.22 &2.44  \\
     StyleNet~\cite{gan2017stylenet}&Humor& 0.010 &0.005  & 99.8\% & 0.024& 0.001&0.12 &0.15 \\
     Random             & Humor& 0.007 & 0.002  & 100\% & 0.006 &0.014 &0.11 &0.07 \\
     DLN-RNN                  & Humor &0.093 & 0.038 & 89.5\% &  0.095 &0.12 &1.58 &0.92 \\  
     DLN                      &Humor &\textbf{0.173} &\textbf{0.065} & 70.0\%&\textbf{0.205} &\textbf{0.182} &\textbf{2.32}&\textbf{2.99} \\
    \midrule
     NST~\cite{kiros2015skip} & Fairy tale& 0.116 &0.044   & 99.8\%&0.116 &\textbf{0.145} &\textbf{2.47} &\textbf{2.44} \\
     StyleNet~\cite{gan2017stylenet}  &Fairy tale& 0.028 &0.013 & 99.8\%& 0.045& 0.026& 0.34&0.46\\
     Random              & Fairy tale & 0.004 & 0.001  & 100\% &0.003 &0.010 &0.06 &0.04\\
     DLN-RNN                  & Fairy tale & 0.084 & 0.033  & 79.5\% & 0.076 &0.140 &1.22 &0.72 \\
     DLN                      &Fairy tale &\textbf{0.135} & \textbf{0.050} & 93.7\%&\textbf{0.194} &0.125 &1.29&2.06\\
    \bottomrule
  \end{tabular}
  \caption{Performance comparison between DLN and several baselines. CS, S and T stand for content similarity, SPICE and transfer accuracy. $p$ and $r$ are as defined in Eq.~\ref{eq:8}. $n_{p}$ and $n_{r}$ are the numerator of each. DLN has generally higher score of content related metrics. Higher is better for all metrics except the transfer accuracy.}
    \label{tab:exp1}
\end{table*}
\subsection{Results} The result of the first experiment is summarized in Table~\ref{tab:exp1}. We also report $p$, $r$ and the numerator of each for further comparison. It is worth noting that the perfect transfer accuracy may not be the best since the model could greedily generate the vocabulary used in the target domain and digress from the image content. Therefore, an ideal stylish description is the one with the high content similarity score and an acceptable transfer accuracy. Our DLN consistently outperforms other baselines in term of all semantic related metrics with a marginal drop of transfer accuracy on most datasets. All baselines are better than Random, which suggests all baselines can generate semantic-related description to certain degree. We observe NST has large $n_{p}$ and $n_{r}$ in fairy tale. We think this is because NST tends to generate long sentences. For each style (Fairy, Humor, Romance, and Lyrics), the average sentence length of NST is $(119,109,103,84)$ while that of DLN is $(38,54,41,97)$. Therefore, it is possible that NST generates more nouns in the unstylish ground truth.

We also report the performance of DLN and DLN-RNN on unstylish description generation task in Table~\ref{tab:exp1-1}. We calculate the BLEU-4, METEOR and CIDEr scores between generated sentences and unstylished ground truth. Combined with the result of stylish description generation in Table~\ref{tab:exp1}, we can conclude that the proposed domain layer norm can benefit the unpaired image to stylish description as we have a better model in conventional image to text generation.

\begin{figure}
    \includegraphics[width=0.40\textwidth]{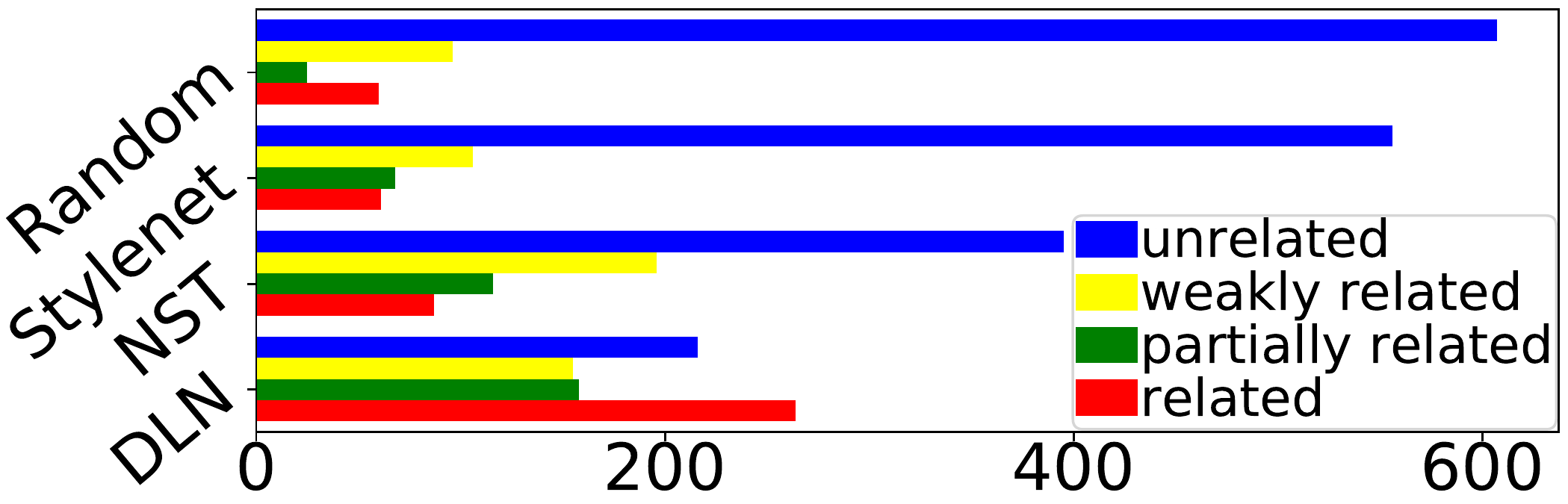}
    \caption{Human study of semantic relevance of all methods. DLN has highest related and lowest unrelated votes compared to other baselines.}
    \label{fig:exp1_human_study}
\end{figure}

\begin{figure*}[t!]
    \includegraphics[width=0.97\textwidth]{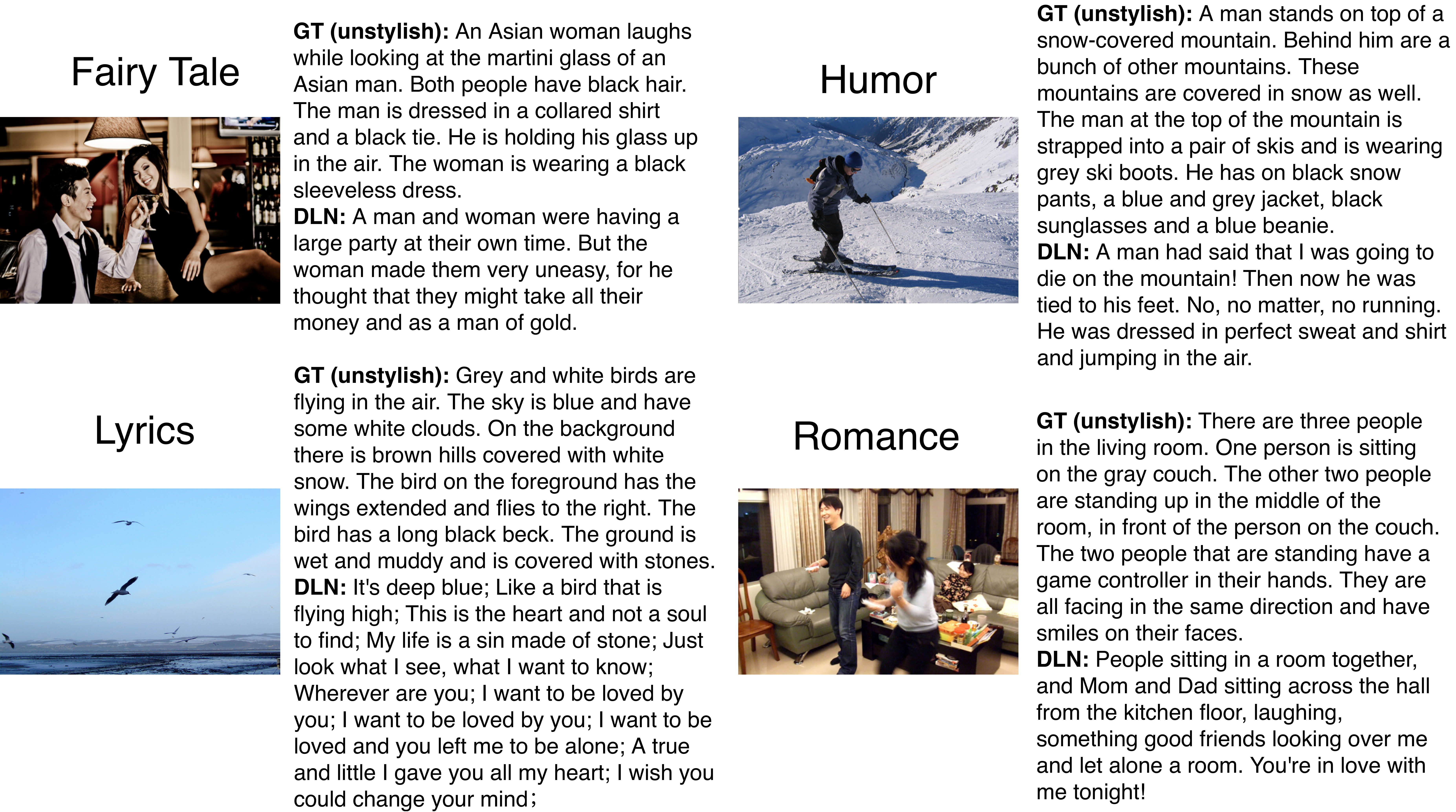}
    \caption{Examples of stylish descriptions by DLN. Note the goal of stylish description is not to match every factual aspect of the image. It should be better judged whether the descriptions are related to the image if the image appears in the context of the target corpus. The semicolon (;) in lyrics serves as new line symbol.}
    \label{fig:exp1_example}
\end{figure*}

The result of human study is shown in Fig~\ref{fig:exp1_human_study}, we report the best of our model in Table~\ref{tab:exp1} (DLN) and other baselines for comparison. The DLN has the highest related and lowest unrelated votes while over half of descriptions are voted as unrelated in other baselines. Qualitative results in Fig~\ref{fig:exp1_example} shows that the description generated by DLN is related to images. Note that the goal of generated stylish description is not to match every factual aspect of images, it should better be judged whether the description is related to the image if the image appears in the target corpus.
\begin{table}[t!]
  \begin{tabular}{lcccc}
    \toprule
    Model  & BLEU-3 & BLEU-4 & METEOR & CIDEr\\
    \midrule
     DLN-RNN&  0.106 & 0.062  & 0.130 & 0.069 \\    
     DLN    &  \textbf{0.132} & \textbf{0.080}  & \textbf{0.150} & \textbf{0.127} \\
    \bottomrule
  \end{tabular}
  \caption{Performance on generate unstylish description. DLN is better than DLN-RNN in all metrics.}
    \label{tab:exp1-1}
\end{table}

\noindent\textbf{Multi-style.}
We progressively expand DLN to include three target domains (fairy, romance, lyrics) to demonstrate the flexibility of our model. In other words, we follow Eq~\ref{eq:7} to train source and fairy tale style then include romance and lyrics style, which is denoted as DLN-Multi. To generate the description, we use the same target decoder with a different style-specific embedding matrix, layer norm parameters, and output matrix. We conduct another human study by asking five workers to determine the best description given following priorities: content, style, and naturalness. This prompt forces workers to choose the better one if the two options are equally related to images. We sample 100 images for each and use the same criteria to select workers. The result is presented in Table~\ref{tab:exp2}, which shows the performance of DLN-Multi is competitive to DLN. DLN-Multi thus gives users the capability to include new style into the existing model, which is a novel feature not reported in other baselines.\\
\noindent\textbf{Discussion: transfer accuracy and domain shift.} We observe a drop in transfer accuracy on the source to humor transfer in DLN, and we believe this is related to the scale of domain shift. To quantify this, we analyze the percentage of shared noun between the source ($V_{src} = 6.2\textrm{k}$) and target domain, which are $(50\%, 68\%, 74\%, 60\%)$ for lyrics, romance humor and fairy tale. For the transfer from the source to humor domain, the shared nouns account for over 70\% nouns in the source domain, which means the domain shift between the source and humor is smaller than others. This makes it more difficult for the classifier to distinguish two domains. Therefore, the transfer accuracy of the source to humor is lower. We note Random get lowest transfer accuracy in lyrics style and we believe this is because sampling word from the vocabulary of lyrics alone cannot have sentences with new line symbol (i.e. ;), which is an important feature for being classified as stylish.
\begin{table}[t!]
  \centering
  \begin{tabular}{lc|cccc}
    \toprule
    Model  & Style & CS& S &T & P\\
    \midrule
     DLN-Multi & Romance&0.116& 0.047 & 97.1\% & 36.7\%\\
     DLN  &Romance &\textbf{0.151}&0.058 & 95.4\% & \textbf{63.3}\%\\
     \midrule
     DLN-Multi & Lyrics& \textbf{0.118}& 0.047  & 99.7\% & \textbf{54.3}\%\\
     DLN  &Lyrics &0.083& 0.033 &99.2\% &  45.8\%\\
    \midrule
     DLN-Multi & Fairy tale&0.120& 0.048 & 99.0\% & 47.4\%\\
     DLN  &Fairy tale &\textbf{0.135}&0.050 & 93.7\% & \textbf{52.6}\%\\
    \bottomrule
  \end{tabular}
  \caption{Result of DLN and DLN-Multi. CS, S, T and P are content similarity, SPICE, tansfer accuracy and human preference score. Overall, the performance of DLN-Multi is competitive to DLN in all metrics.}
    \label{tab:exp2}
\end{table}

\section{Conclusion and future work}
We propose a novel unsupervised stylish IDG model via domain layer norm with the capability to progressively include new styles. Experiment results show that our stylish IDG results are more preferred by human subjects. We plan to invesitgate the intermediate style generated by interpolation of domain layer norm parameter and address the fluency of generated sentences in the future.
\small{
\bibliography{main}
\bibliographystyle{aaai}
}
\clearpage
\section{Appendix}
The content of this supplementary material is summarized as below:
\begin{itemize}
    \item Data statistic
    \item Implementation details of baselines and DLN
    \item Interface of human evaluation
    \item More qualitative examples
\end{itemize}
\section{Data Statistic}
\begin{figure*}[t!]
    \centering
    \includegraphics[width=0.6\textwidth]{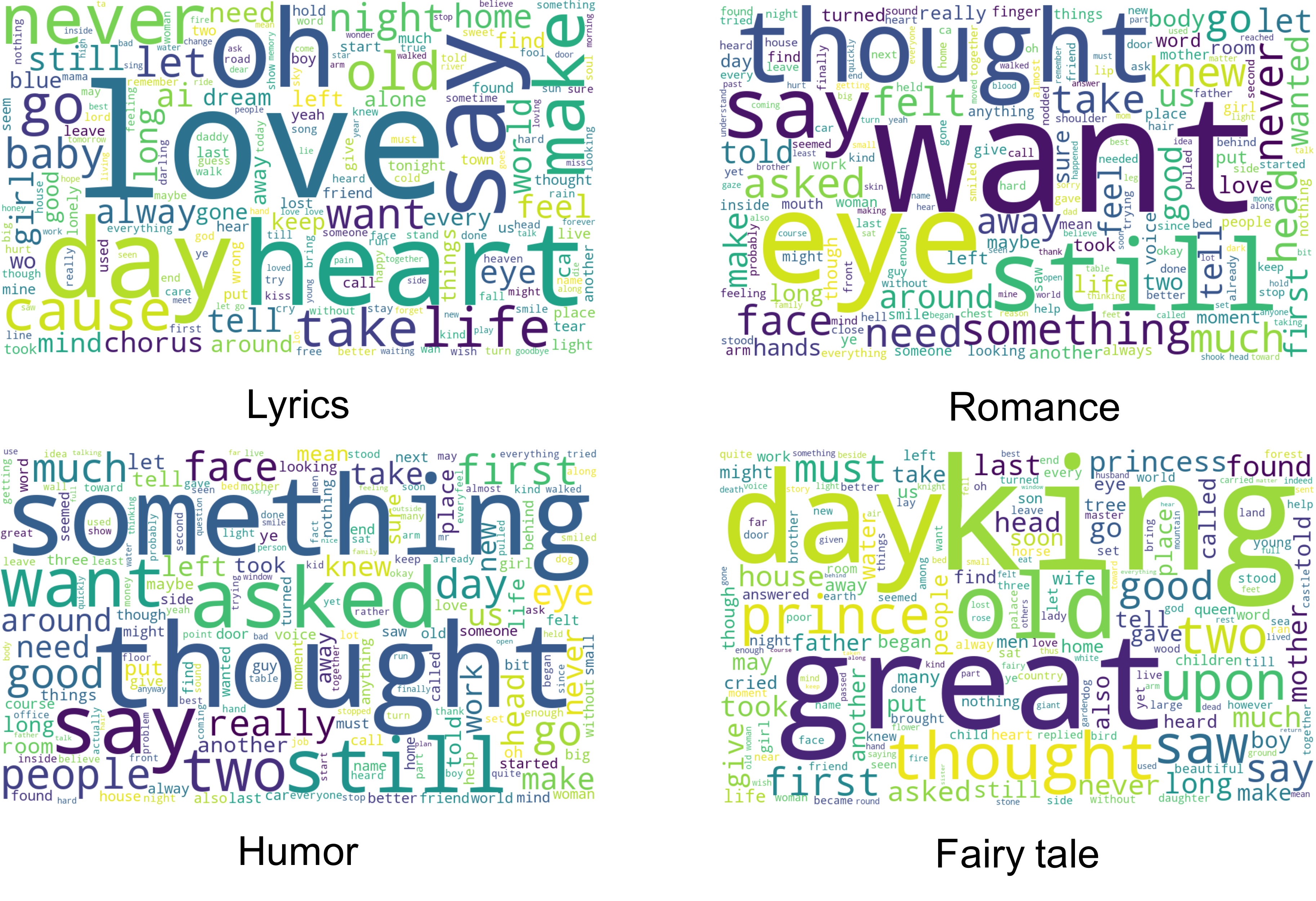}
    \caption{Word cloud of our four datasets}
    \label{fig:wordcloud}
\end{figure*}

For romance and humor data, we randomly sampled 50000 passages from BookCorpus~\cite{zhu2015aligning}. For fairy tale, we crawled from the website\footnote{http://www.loyalbooks.com} and also sample 50000 passages from it. For the country song lyrics, we use the data released by Kaggle\footnote{https://www.kaggle.com/gyani95/380000-lyrics-from-metrolyrics} and use all country song lyrics as our corpus. The Fig~\ref{fig:wordcloud} shows the word usage of each corpus.

\section{Baselines}
\begin{itemize}
    \item \textbf{Neural Story Teller}~\cite{kiros2015skip}: NST contains two separate modules: image to caption in the source domain and stylish story decoder in the target domain. The first module is used to extract the source domain textual representation of the image. They use pre-trained image caption alignment~\cite{kiros2014unifying} on MS-COCO to extract the top $N$ neighbor captions for an image. The textual representation of the image is calculated by averaging the skip-thought encoded vector of top $N$ captions. The second module is used to reconstruct target domain textual representation to original stylish story passage, where a decoder is trained to reconstruct the textual representation of stylish story passage to the original stylish story passage. The textual representation here is also the skip-thought encoded vector. To generate the description for given image, it first subtracts the mean of all skip-thought encoded MS-COCO caption and added the mean of skip-thought encoded stylish story passage to transform the source textual representation to the target one; then, feed it into the second module to generate the stylish story.
    
    \item \textbf{StyleNet}~\cite{gan2017stylenet}: We re-implement StyleNet as one of our baseline. The StyleNet is based on factorized LSTM, which factorized the weights mapping inputs to hidden representation to $\boldsymbol{\theta_W} = \boldsymbol{U}\boldsymbol{S}\boldsymbol{V}$, where $\boldsymbol{U} \in \mathbb{R}^{p_e \times p_m}$, $\boldsymbol{S} \in \mathbb{R}^{p_m \times p_n}$ and $\boldsymbol{V} \in \mathbb{R}^{p_n \times p_h}$, where $p_e$,$p_h$ are dimension of embedding and hidden size, and $p_m$,$p_n$ are the dimension of the matrix. It contains a language model trained on target corpus and a source image to caption decoder where each factorized LSTM is associated with a style matrix $\boldsymbol{S}_{s}$ and $\boldsymbol{S}_{t}$. During inference, it replace the source style matrix with target style matrix and generate descriptions.
\end{itemize}

\section{Implementation Details}
\noindent\textbf{NST.} We follow the original setting and implementation\footnote{https://github.com/ryankiros/neural-storyteller} and train the decoder on our corpus till converge. We set the maximum length of text data equals to 100.

\noindent\textbf{DLN.} We use the same text length as used in NST. We use most common 10000 vocabularies in the source domain. For target domain, we use 10000 vocabularies for lyrics and 15000 vocabularies for romance, humor and fairy tale corpora.

We use skips-thought vector released by Tensorflow~\cite{abadi2016tensorflow} as our $E_{T}$. We follow the original NST implementation to stack uni-skip and bi-skip skips-thought vector to get 4800 dimension feature for text. We use pre-split training set in VG-Para~\cite{krause2016paragraphs}.

We use ResNet50 from Keras\footnote{https://keras.io/applications/} as our $E_{I}$. The dimension of our latent space is 620, which is the same as our word embedding dimension. In implementation, we fix $E_{I}$ and $E_{T}$ and append projection matrix $\boldsymbol{\theta}_{P_I}$ and $\boldsymbol{\theta}_{P_T}$ as the last layers. During training, we only update projection matrix. We initialize all weight matrix by uniform initialization. The number of hidden units used in LN-LSTM is 1000, and we optimize our model by Adam optimizer~\cite{kingma2014adam} with start learning rate as 0.001 and decayed factor as 0.5 every 80. We follow NST to use gradient clipping = 5 in DLN. The training epoch is 100 with batch size as 64. We choose $\lambda=0.5$ for DLN training. 

\noindent\textbf{DLN-Multi} The vocabulary size of each domain is 5500. We choose $\lambda_1=0.2$ and $\lambda_2=0.1$. Other settings and hyperparamters are the same as DLN except the regularization term 
\begin{align}
R=\lVert\boldsymbol{\theta}^{\prime}_{\boldsymbol{W}} - \boldsymbol{\theta_{W}} \rVert_{2} + \lVert \boldsymbol{\theta}^{\prime}_{\boldsymbol{V}} - \boldsymbol{\theta_{V}} \rVert_{2} +
\lVert \boldsymbol{\theta}^{\prime}_{E_{I}} - \boldsymbol{\theta}_{E_{I}}\rVert_{2}
\end{align}
We implement the $\boldsymbol{\theta}^{\prime}_{\boldsymbol{W}}$ by $\boldsymbol{\theta}^{\prime}_{\boldsymbol{W}} = \boldsymbol{\theta_W} || \boldsymbol{\theta_{\Delta W}}$, where $||$ is the concatenation operation of matrix and $\boldsymbol{\theta_{\Delta W}}$ is the new vocabulary used in the new style. In the subtraction, we only subtract the $\boldsymbol{\theta_W}$ part in $\boldsymbol{\theta}^{\prime}_{\boldsymbol{W}}$ to match the dimension of matrix. Similar logic can be applied to $\boldsymbol{\theta}^{\prime}_{\boldsymbol{V}}$. We use the projection weight $\boldsymbol{\theta}_{P_I}$ learned during pre-training and the weight during the training of DLN-Mutli as our $\boldsymbol{\theta}^{\prime}_{E_{I}}$ and  $\boldsymbol{\theta}_{E_{I}}$.

\begin{figure*}[h!]
    \centering
    \includegraphics[width=0.6\textwidth]{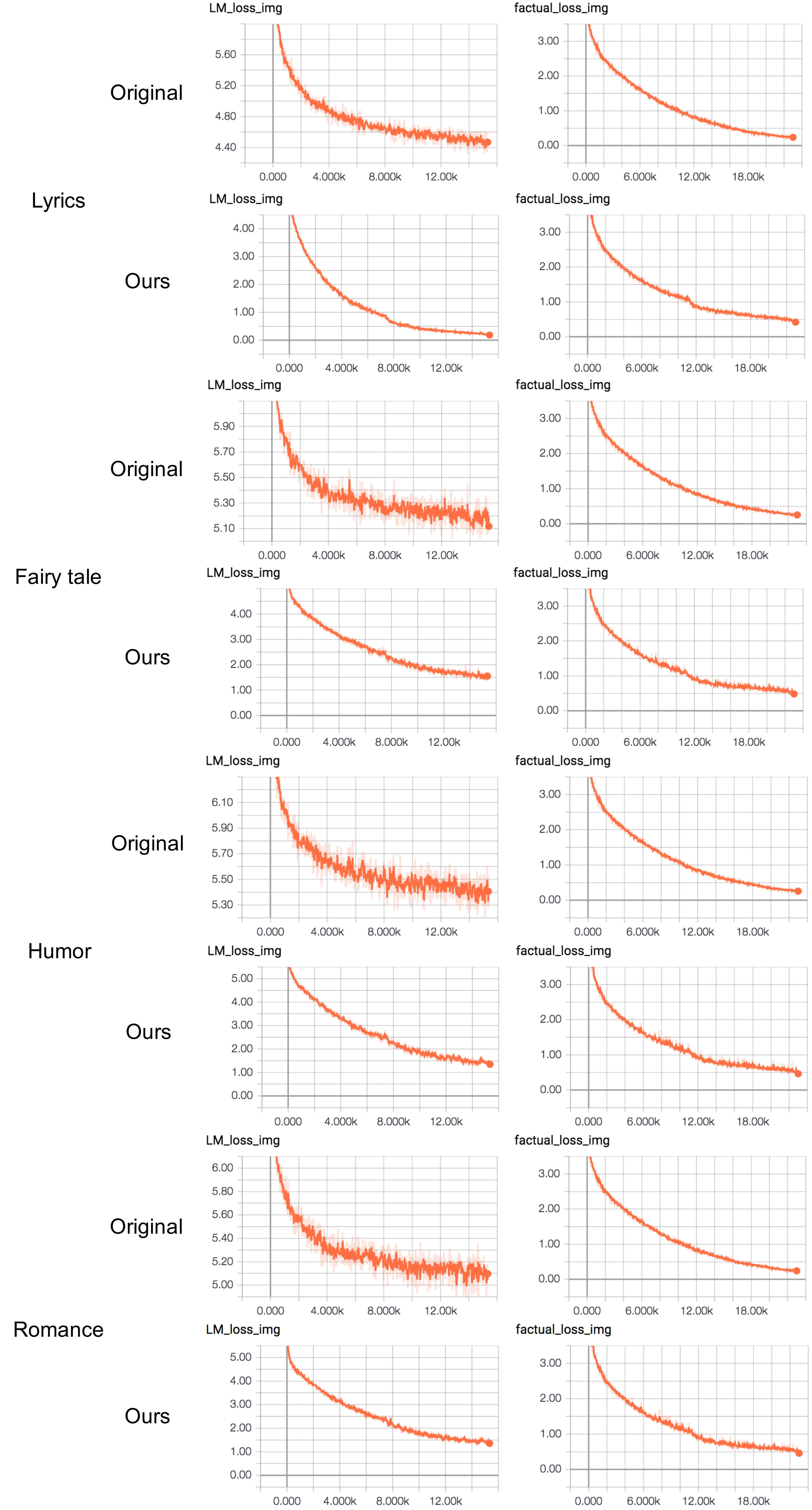}
    \caption{Comparison of training StyleNet by method mentioned in original paper (Original) and our modification (Ours). The LM\_loss and factual\_img\_loss refer to $\mathcal{L_{T}}$ and $\mathcal{L_{S}}$ respectively}
    \label{fig:loss_comparison}
\end{figure*}

\noindent\textbf{StyleNet} To train StyleNet, we use the same hidden unit of LSTM as DLN and follow the iterative training method reported in the original paper except that we also update the share weight when training language model on target corpus, which we found this modification has better convergence in our task as shown in Fig~\ref{fig:loss_comparison}. We also apply our decay learning setting, which we found has faster convergence. During inference, we follow StyleNet and NST by using beam search with a beam width of 5 and the unknown token for all methods.

\noindent\textbf{Pretrained classifier in transfer accuracy.} For the classifier used in evaluation, we use convolutional network proposed in~\cite{kim2014convolutional}. We train the classifier to achieve over 99\% accuracy to distinguish $\mathcal{D}_S$ and $\mathcal{D}_T$.
\section{Human Evaluation Setup }
\begin{figure*}[h!]
    \centering
    \includegraphics[width=0.8\textwidth]{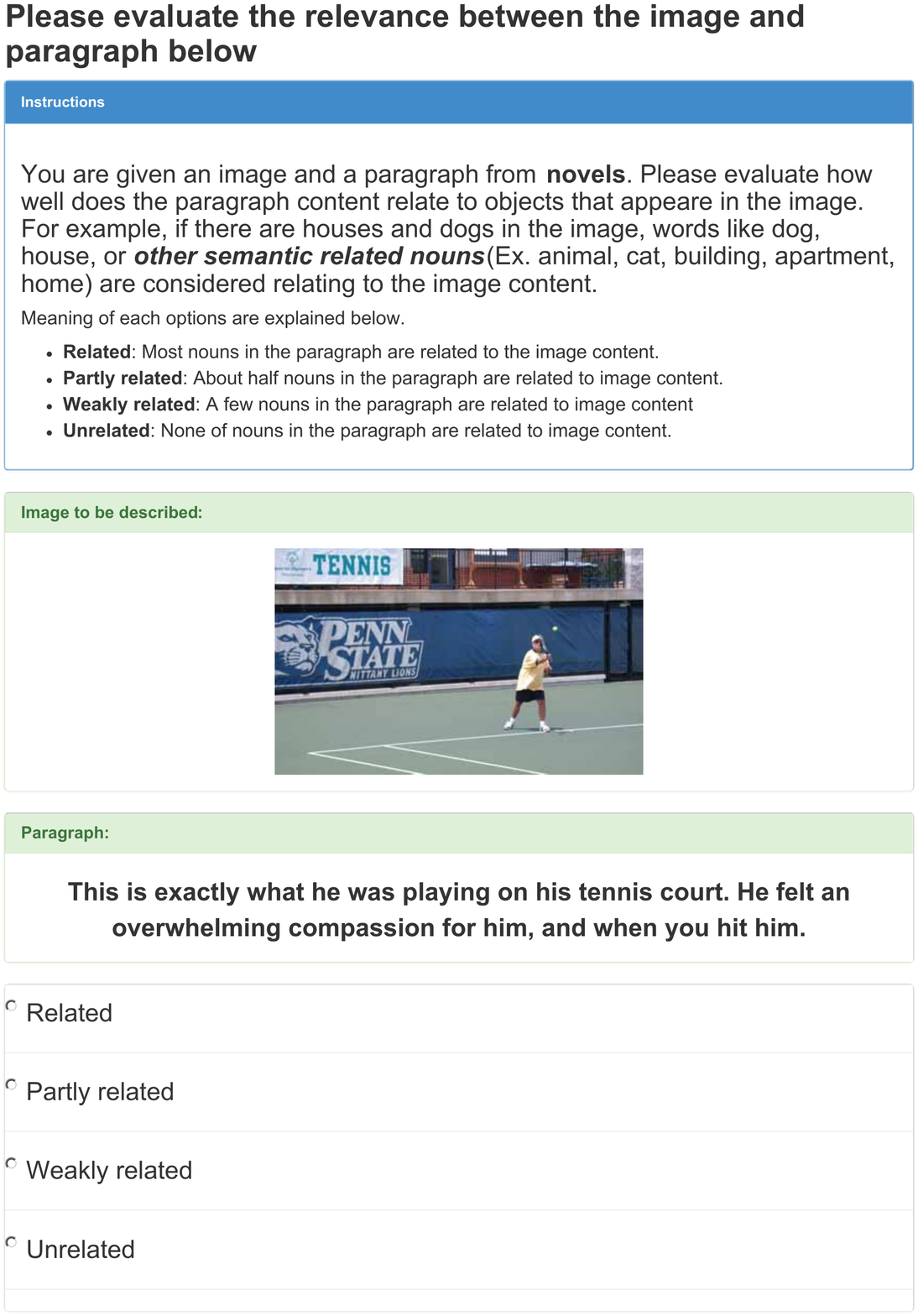}
    \caption{A screen-shot of human study on semantic relevance of generated paragraphs.}
    \label{fig:human_study_1}
\end{figure*}

We performed two human evaluation tasks using the Amazon Mechanical Turk\footnote{https://www.mturk.com} platform. The first was a relevance task, asking how well does descriptions relate to the image content on a four level scale. We provide screen-shots of the instructions given to workers in Fig~\ref{fig:human_study_1}. The second study aims to compare the attractiveness of descriptions generated by the DLN mode and DLN-Multi. Fig~\ref{fig:human_study_1} is the screen-shots given to works for this experiment.
To ensure reliable results and avoid workers who choose randomly, only workers with more than 70\% accuracy and 100 successful HITs previously are allowed to attend the study. In the second human study, we also provide a dummy text as trap option to monitor the labelling quality. The result shows almost no trap options are chosen, indicating experiment results to be reliable.

\begin{figure*}[h!]
    \centering
    \includegraphics[width=0.8\textwidth]{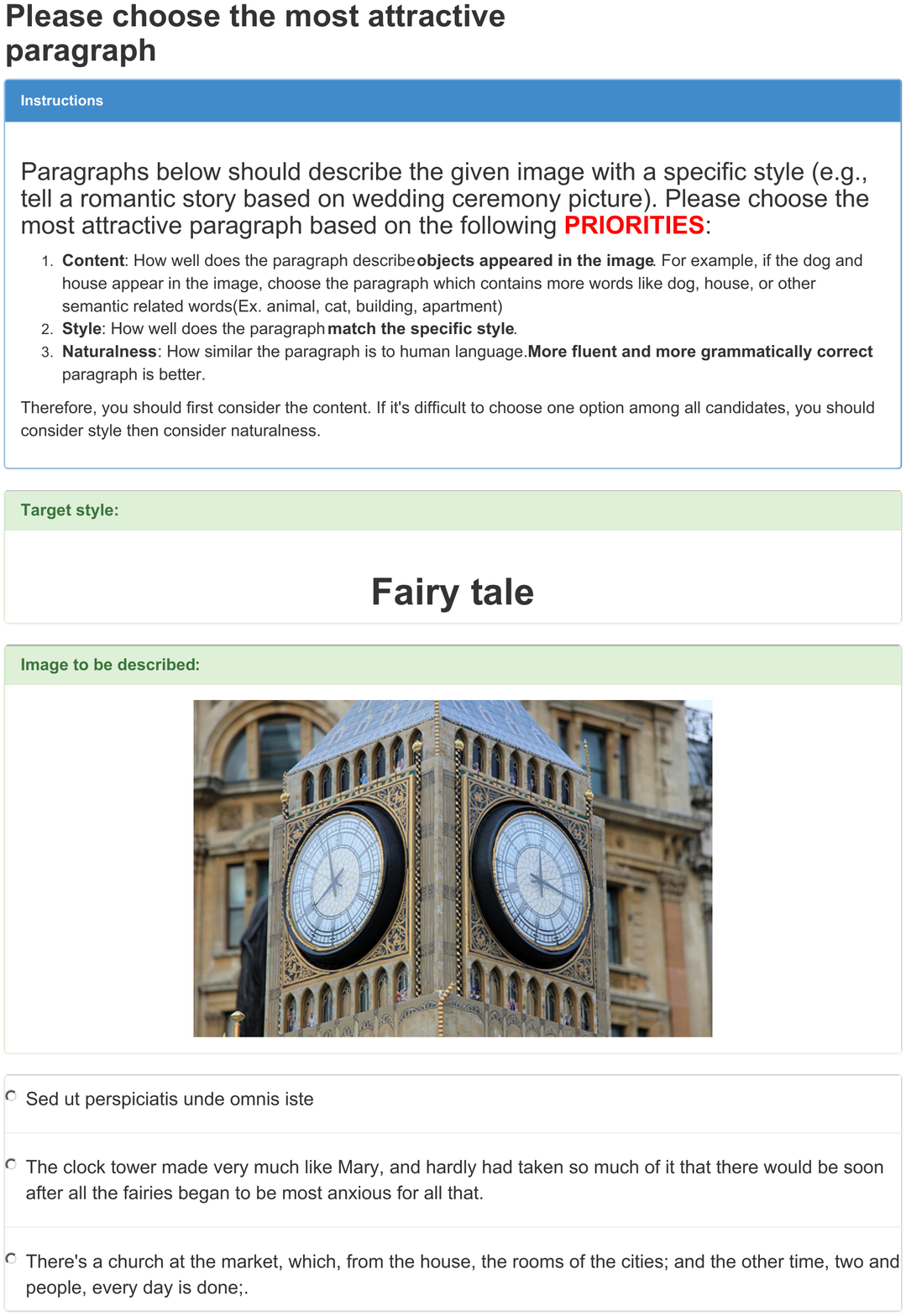}
    \caption{A screen-shot of human study on paragraph attractiveness.}
    \label{fig:human_study_2}
\end{figure*}

\section{Qualitative example of stylish image description generation}
\label{sec:example_naive}
\begin{figure*}[t!]
    \centering
    \includegraphics[width=0.99\textwidth]{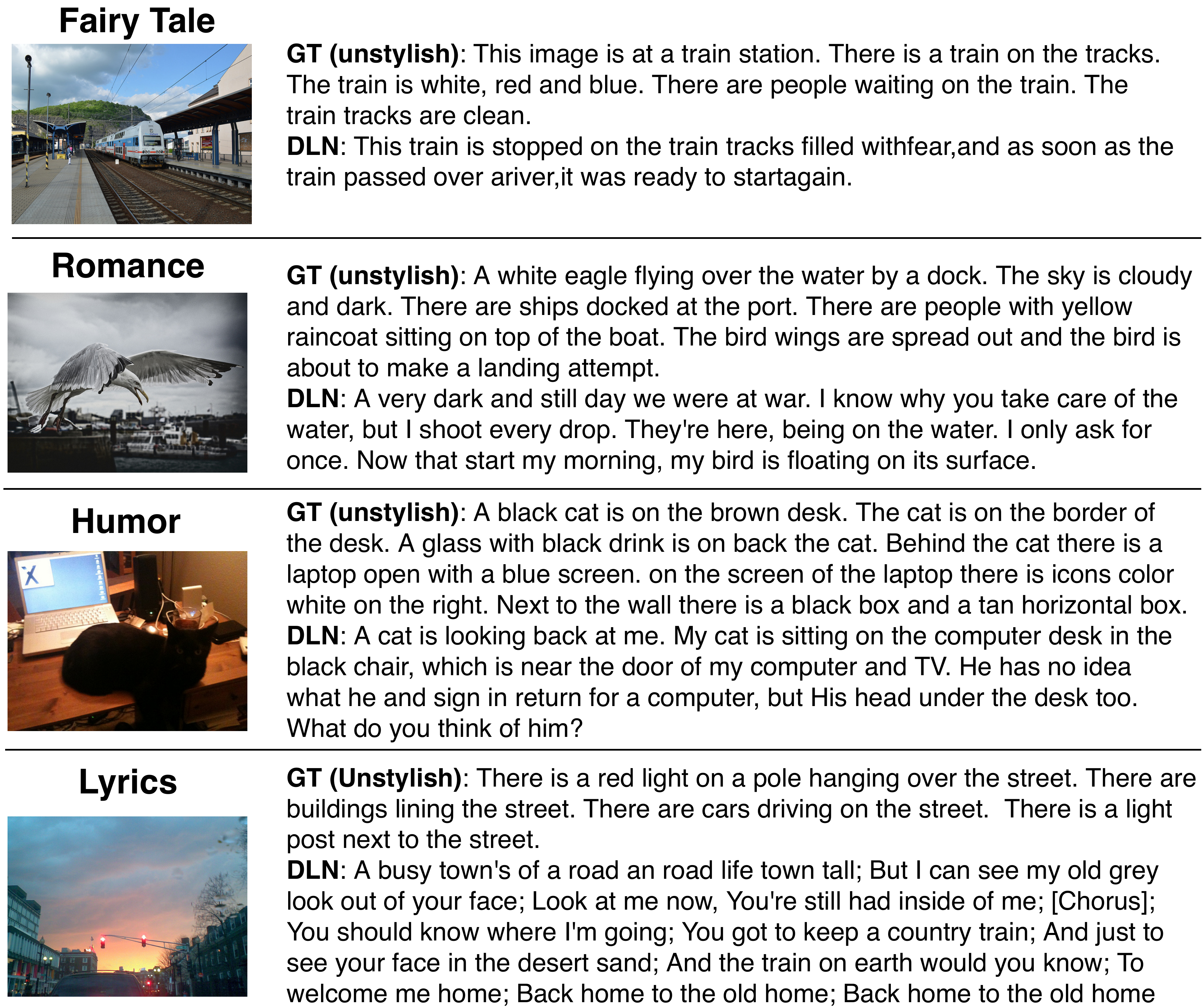}
    \caption{Examples of stylistic image description by DLN.}
    \label{fig:supp_exp1}
\end{figure*}

We demonstrate more qualitative examples of stylish image description generated by DLN in Fig~\ref{fig:supp_exp1}.
\end{document}